\documentclass[11pt]{article}

\PassOptionsToPackage{hyperfootnotes=false}{hyperref}
\usepackage[preprint]{acl}

\usepackage{times}
\usepackage{latexsym}

\usepackage[T1]{fontenc}

\usepackage[utf8]{inputenc}

\usepackage{microtype}

\usepackage{inconsolata}

\usepackage{graphicx}
\usepackage{booktabs}

\title{\textsc{Stage}: Stateful Translation to Agentic Graph Execution with Policy-Scoped Context and Deterministic Control}

\author{
  Mengxi Luo\thanks{These authors contributed equally.} \quad Changjia Chen\footnotemark[1] \quad An Cao \quad Wanyi Dai \quad Zirong Huang\\
  \textbf{Shixiong Gao \quad Ruoqi Shu \quad John Chen \quad Bo Wan} \\[0.5em]
  BMO Financial Group \\
  \texttt{mengxi.luo@bmo.com, changjia.chen@bmo.com} \\
  \texttt{an.cao@bmo.com,  wanyi.dai@bmo.com, zirongvera.huang@bmo.com}\\
  \texttt{shixiong.gao@bmo.com, ruoqi.shu@bmo.com, john4.chen@bmo.com, bo.wan@bmo.com}\\
  \small{
   \textbf{Correspondence:} \href{mailto:changjia.chen@bmo.com}{changjia.chen@bmo.com}
  }
}

\begin{document}
\maketitle


\begin{abstract}

Enterprise agents are expected to deliver correct and consistent outcomes while executing complex procedures, adhering to governing policies, and operating within practical resource budgets. We present \textsc{Stage}, an integrated execution framework combining typed task nodes, shared execution memory, document tools, and deterministic orchestration. We evaluate \textsc{Stage} on SOP-Bench Referral Abuse, the Retail and Telecom domains of $\tau^2$-bench, and two proprietary banking workflows, Smart Dispute and Paystub Fraud Detection. Using the same underlying model in each comparison, \textsc{Stage} improves single-run success and repeated-run reliability over the baseline in all 18 benchmark--model pairs. Success across all three runs (\(\mathrm{Pass}^{3}\)) increases by 7.5--66.0 percentage points. Fixed-graph ablations demonstrate the contributions of localized policy context and coordinator-owned routing. Further analysis of Smart Dispute shows improved coverage of required operations across procedure lengths. These findings show that integrating local reasoning, persistent state, and procedural control can improve reliable enterprise workflow completion without changing the underlying model.

\end{abstract}

\section{Introduction}
\label{sec:introduction}

Enterprise deployment places several simultaneous demands on LLM agents, including correct and consistent task completion, acceptable operating cost, and procedural compliance with the governing policy. Enterprise workflows connect local judgments through conditional rules and facts established earlier in a procedure \citep{yang2026business,liu2026analyzing,singh2026agentops,zhang2026finreporting}. An LLM agent can therefore make a correct local judgment yet fail to complete the business process that depends on it. The execution system must achieve high accuracy while preserving relevant information across steps and ensuring that required actions follow the conditions and dependencies specified by policy. Meeting these obligations within practical resource budgets makes the organization of the execution system central to effective enterprise deployment.

A monolithic agent must interpret policy and case evidence while retaining relevant facts, tracking procedural progress, and selecting subsequent actions. These responsibilities place competing demands on the same reasoning process as workflow complexity and accumulated context increase \citep{liu2024lost}. Recent work on workflow compilation and graph-based agent systems highlights the value of making data dependencies, execution state, and control flow explicit \citep{xu2026artic,feng2026graphengineering}. Building on this direction, we ask \textit{how an LLM execution system should be designed to reliably carry out complex enterprise workflows at an acceptable cost while maintaining compliance}. Localizing policy reasoning requires the execution system to preserve information and procedural dependencies across task boundaries. We approach this question through the integrated design of bounded reasoning tasks, shared execution state, evidence access, and runtime coordination.

\begin{figure*}[t]
\centering
\includegraphics[width=\textwidth]{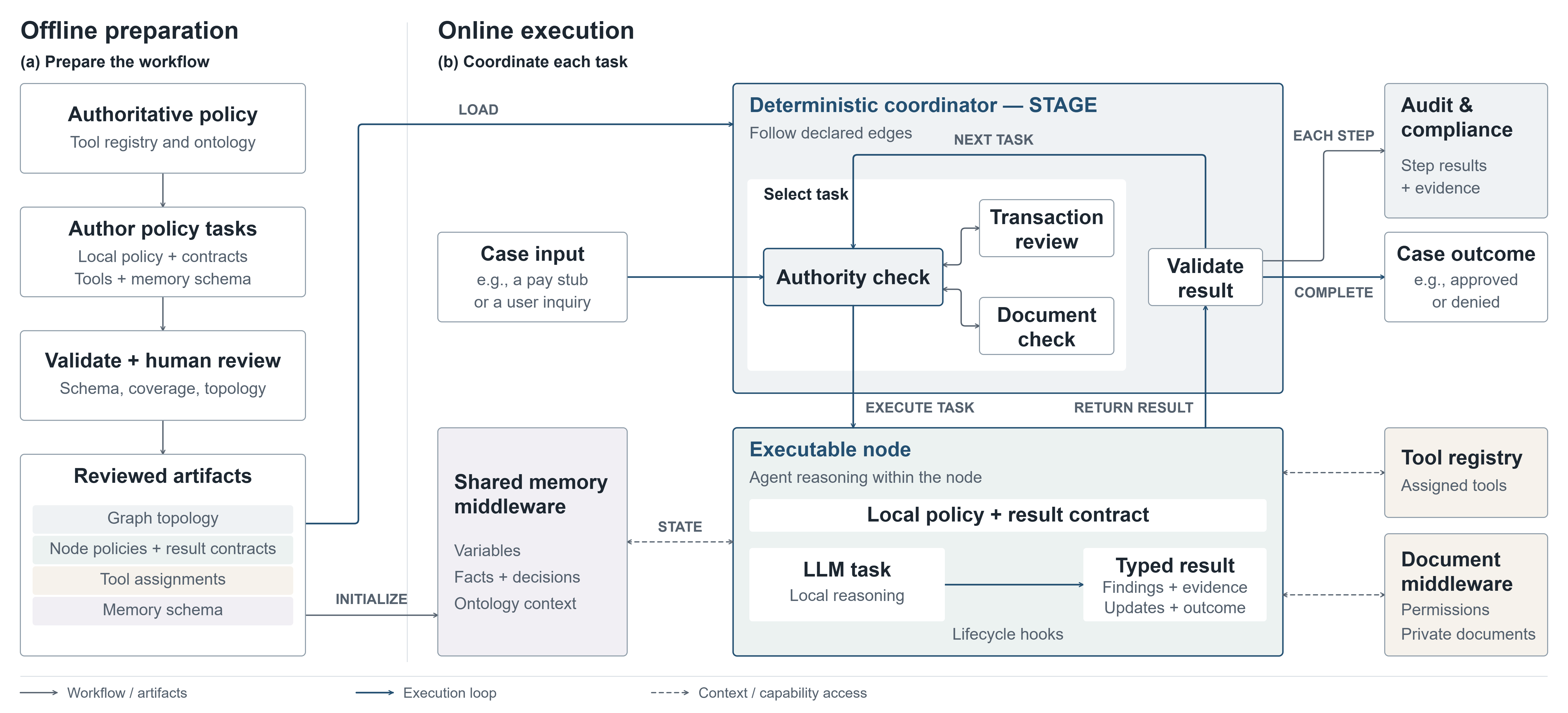}
\caption{\textsc{Stage} overview. Offline preparation produces reviewed workflow artifacts (left). Online execution combines deterministic coordination with policy-scoped agent reasoning, shared memory, and controlled tools and document access; step results and evidence support audit and compliance review (right).}
\label{fig:dispute-comparison}
\end{figure*}

We introduce \textsc{Stage}, an integrated execution framework for enterprise workflows that is being evaluated in an internal banking pilot. It translates authoritative workflow instructions into a reviewed executable graph that connects bounded policy reasoning through persistent case state (Figure~\ref{fig:dispute-comparison}). The runtime controls procedural progression, allowing each agent to focus on its current task. This division of responsibility follows the principle that \textit{the LLM owns local meaning while code owns control}.

Our contributions are threefold. First, we present an integrated execution design for enterprise workflows under policy and resource constraints. Second, we evaluate task success, repeated-run reliability, and inference cost across public benchmarks and proprietary banking workflows using matched underlying models. Third, we isolate policy localization and routing ownership through fixed-graph ablations and analyze procedural coverage across workflow lengths. These comparisons characterize the reliability benefits and inference-cost tradeoffs of the design.

\section{\textsc{Stage}}
\label{sec:framework}

\paragraph{Executable Policy Graph.}
\label{sec:typed-graph}

\textsc{Stage} represents authoritative workflow instructions as an executable graph. The graph is constructed offline through an LLM-assisted procedure that decomposes the policy into bounded tasks and assigns each node task-relevant context and appropriate tools from the tool registry. The candidate graph and these assignments are validated and reviewed before being frozen for execution and evaluation. This separation makes the procedural structure explicit before individual cases are processed and provides a common basis for controlled comparisons. The construction procedure is detailed in Appendix~\ref{sec:policy2graph-construction}.

\paragraph{Compliance-Carrying Execution Nodes.}
\label{sec:node-execution}

Each node carries the policy obligations for a bounded task together with explicit conditions for accepting its result. Its prompt specifies the required checks and decision criteria, while the assigned tools constrain the operations available during execution. Typed result contracts specify findings and supporting evidence. Directed edges restrict progression to permitted successors, with decision branches selected only from declared outcomes. Lifecycle hooks support additional policy checks that can be evaluated deterministically. Compliance requirements therefore accompany each task through execution, with encoded constraints enforced by the runtime and semantic judgments delegated to the model.

\paragraph{Shared Execution State and Evidence Access.}

\textsc{Stage} separates persistent task state from transient node context. Shared memory represents reusable facts and intermediate decisions as structured, scoped variables, giving nodes access to case state and relevant enterprise ontology context. The runtime injects a memory snapshot into each node and applies updates through its typed result. Document middleware provides guardrails for enterprise use by gating access to private information through configurable document permissions and limiting repeated retrieval. Together, these mechanisms preserve information across task boundaries and support evidence reuse while allowing individual nodes to reason within a focused context.

\paragraph{Coordinator-Owned Runtime Control.}
\label{sec:coordination}

The coordinator executes the reviewed workflow graph and owns procedural progression. At each step, it provides the node contract and current case state to the configured executor, validates the returned result against its contract, and applies accepted memory updates and declared transitions. Models and agent types can be selected per node through a common execution interface. This design allows task-specific execution strategies within a shared control structure, separating how a task is solved from how the workflow progresses.

\paragraph{Research Questions.}
Having defined the \textsc{Stage} execution model, we next examine its empirical consequences through three research questions.

\textbf{RQ1:}
Does an integrated execution architecture centered on localized policy reasoning improve task success and repeated-run reliability across enterprise workflows and underlying models?

\textbf{RQ2:}
Holding the reviewed graph fixed, how do policy localization and routing ownership affect execution reliability?

\textbf{RQ3:}
How well does \textsc{Stage} maintain compliance with governing procedures as procedure length increases, compared with the baseline?

\section{Experimental Settings}
\label{sec:experimental-settings}

\subsection{Benchmarks and Models}
\label{sec:setup}

We evaluate \textsc{Stage} on five workflows spanning public benchmarks and proprietary banking procedures with different levels of procedural complexity. SOP-Bench Referral Abuse tests policy-guided investigation and decision-making with executable tools \citep{nandi2025sopbench}. Retail and Telecom from \(\tau^2\)-bench represent shallower and deeper workflows, respectively \citep{barres2025tau2bench}. Smart Dispute and Paystub Fraud Detection are based on proprietary banking policies and provide deep workflows for dispute handling and paystub verification, respectively. The reconstructed workflow graphs are presented in Appendix~\ref{sec:benchmark-graphs}.

Smart Dispute and Paystub Fraud Detection contain 50 and 51 proprietary cases, respectively. Both benchmarks evaluate success through case-level outcomes, which form the basis of the reported \(\mathrm{Pass}^{1}\) and \(\mathrm{Pass}^{3}\) rates. Smart Dispute's outcome-based evaluation criteria and procedural metrics are detailed in Appendix~\ref{sec:smart-dispute-analysis}.

We evaluate the public workflows with GPT-5.6 Luna, Claude Haiku 4.5, Amazon Nova 2 Lite, and Gemma 4 26B A4B. Model selection prioritizes affordability and inference efficiency, reflecting practical constraints on enterprise deployment. For each model, we compare \textsc{Stage} with the baseline using the same underlying model. Evaluation on the proprietary banking workflows excludes Gemma 4 26B A4B.

\subsection{Experimental Design}

For each workflow, we generate, validate, approve, and freeze one execution graph together with its node-level policy context and tool assignments before evaluation. All conditions use the same DeepAgent scaffold. Within each comparison, we hold the model, task input, available evidence, environment, and evaluator fixed. Public workflows retain their original tools and success criteria, while Smart Dispute and Paystub Fraud Detection use their respective internal evaluation specifications. All reported results use the frozen graphs without post-evaluation manual adjustment.

We use \textbf{baseline} to denote monolithic execution, in which one agent receives the complete policy and manages both reasoning and procedural progression. RQ1 compares this baseline with the integrated \textsc{Stage} architecture. \textsc{Stage} distributes policy reasoning across bounded tasks connected through shared execution state and coordinator-owned control.

RQ2 evaluates two fixed-graph ablations on Telecom and Smart Dispute. \textbf{Full-policy nodes} replace each node's localized policy context with the complete policy while retaining coordinator-owned routing. \textbf{LLM routing} retains node-scoped context but lets the model select any node in the frozen graph, including non-successors. We hold topology, node objectives, case evidence, permitted capabilities, result interfaces, and attempt budgets fixed. Each ablation therefore changes only policy scope or routing ownership. RQ3 stratifies Smart Dispute results by required procedure length, defined as the number of required operations in each case.

We report \(\mathrm{Pass}^{1}\) as the mean case-level success rate across three runs and \(\mathrm{Pass}^{3}\) as the proportion of cases completed successfully in all three runs. We prioritize \(\mathrm{Pass}^{3}\) because it measures correct completion across repeated executions, a key requirement for enterprise workflows. Appendix~\ref{sec:inference-cost} reports agent inference costs. For RQ3, \textbf{dependency coverage} measures the fraction of required operations matched under the evaluator's dependency checks. \textbf{Strict success} requires completed execution with the correct terminal outcome, no forbidden action attempts, and all required operations matched. These metrics distinguish partial procedural progress from successful completion of the full procedure. Appendix~\ref{sec:smart-dispute-analysis} provides the detailed definitions.

\section{Results}
\label{sec:results}

\subsection{RQ1: Overall Effectiveness}
\label{sec:success-results}

\begin{table*}[t]
\centering
\small
\setlength{\tabcolsep}{5pt}
\renewcommand{\arraystretch}{1.08}
\begin{tabular}{@{}llrrr|rrr@{}}
\hline
& & \multicolumn{3}{c|}{\(\mathrm{Pass}^{1}\) (\%)}
& \multicolumn{3}{c}{\(\mathrm{Pass}^{3}\) (\%)} \\
\textbf{Benchmark} & \textbf{Model}
& \textbf{Base} & \textbf{\textsc{Stage}} & \(\mathbf{\Delta}\)
& \textbf{Base} & \textbf{\textsc{Stage}} & \(\mathbf{\Delta}\) \\
\hline
Referral Abuse & GPT-5.6 Luna & 94.3 & \textbf{97.8} & \(+3.5\) & 88.0 & \textbf{95.5} & \(+7.5\) \\
 & Claude Haiku 4.5 & 90.7 & \textbf{96.0} & \(+5.3\) & 79.0 & \textbf{88.5} & \(+9.5\) \\
 & Amazon Nova 2 Lite & 84.8 & \textbf{96.3} & \(+11.5\) & 73.0 & \textbf{89.0} & \(+16.0\) \\
 & Gemma 4 26B A4B & 92.3 & \textbf{98.0} & \(+5.7\) & 84.5 & \textbf{94.5} & \(+10.0\) \\
\hline
Retail & GPT-5.6 Luna & 55.8 & \textbf{73.3} & \(+17.5\) & 37.5 & \textbf{65.0} & \(+27.5\) \\
 & Claude Haiku 4.5 & 51.7 & \textbf{80.8} & \(+29.2\) & 27.5 & \textbf{72.5} & \(+45.0\) \\
 & Amazon Nova 2 Lite & 48.3 & \textbf{75.8} & \(+27.5\) & 32.5 & \textbf{65.0} & \(+32.5\) \\
 & Gemma 4 26B A4B & 78.3 & \textbf{80.8} & \(+2.5\) & 60.0 & \textbf{70.0} & \(+10.0\) \\
\hline
Telecom & GPT-5.6 Luna & 45.8 & \textbf{95.8} & \(+50.0\) & 32.5 & \textbf{92.5} & \(+60.0\) \\
 & Claude Haiku 4.5 & 50.0 & \textbf{90.8} & \(+40.8\) & 25.0 & \textbf{72.5} & \(+47.5\) \\
 & Amazon Nova 2 Lite & 60.8 & \textbf{72.5} & \(+11.7\) & 27.5 & \textbf{57.5} & \(+30.0\) \\
 & Gemma 4 26B A4B & 47.5 & \textbf{91.7} & \(+44.2\) & 27.5 & \textbf{80.0} & \(+52.5\) \\
\hline
Smart Dispute & GPT-5.6 Luna & 70.7 & \textbf{96.0} & \(+25.3\) & 54.0 & \textbf{96.0} & \(+42.0\) \\
 & Claude Haiku 4.5 & 34.7 & \textbf{95.3} & \(+60.7\) & 24.0 & \textbf{90.0} & \(+66.0\) \\
 & Amazon Nova 2 Lite & 43.3 & \textbf{81.3} & \(+38.0\) & 24.0 & \textbf{70.0} & \(+46.0\) \\
\hline
Paystub Fraud Detection & GPT-5.6 Luna & 28.8 & \textbf{79.7} & \(+51.0\) & 21.6 & \textbf{68.6} & \(+47.1\) \\
 & Claude Haiku 4.5 & 27.5 & \textbf{90.8} & \(+63.4\) & 17.6 & \textbf{82.4} & \(+64.7\) \\
 & Amazon Nova 2 Lite & 31.4 & \textbf{81.7} & \(+50.3\) & 27.5 & \textbf{62.7} & \(+35.3\) \\
\hline
\end{tabular}
\caption{Single-run task success and repeated-run reliability. \(\mathrm{Pass}^{1}\) denotes single-run success, while \(\mathrm{Pass}^{3}\) requires successful completion in all three runs. \(\Delta\) is the percentage-point difference between \textsc{Stage} and the baseline.}
\label{tab:success-reliability}
\end{table*}

Table~\ref{tab:success-reliability} compares \textsc{Stage} with the baseline using the same underlying model across five workflows. \textsc{Stage} improves both single-run success and repeated-run reliability in all 18 benchmark--model comparisons.

\paragraph{Public benchmarks.}

On Referral Abuse, \textsc{Stage} improves \(\mathrm{Pass}^{3}\) by 7.5--16.0 percentage points despite relatively strong baseline performance. Retail shows gains of 10.0--45.0 points across models. Telecom, which requires deeper routing and coordination between agent and user actions, shows improvements of 30.0--60.0 points.

\paragraph{Banking workflows.}
\textsc{Stage} also improves repeated-run success across both proprietary banking workflows. On Smart Dispute, \(\mathrm{Pass}^{3}\) increases by 42.0--66.0 percentage points, reaching 70.0--96.0\% across the three models. Paystub Fraud Detection shows gains of 35.3--64.7 points.

Across public and banking workflows, these consistent gains support structured execution as a means of improving the reliable delivery of correct case-level outcomes under each benchmark's evaluation criteria.

\paragraph{Inference cost.}
Agent inference costs estimated from usage recorded during the RQ1 experiments are reported in Table~\ref{tab:inference-cost} and discussed in Appendix~\ref{sec:inference-cost}. Structured execution does not invariably increase cost. \textsc{Stage} costs less than the baseline for Gemma on Retail and Nova on Smart Dispute, although most model--workflow pairs incur additional cost. Efficiency depends on how readily the model handles the workflow and satisfies node-level requirements. Additional turns and retries, together with structured outputs for compliance checks and review, can increase inference cost. Overall, we consider the reported agent inference costs acceptable relative to the observed reliability gains in these experiments.

\begin{figure*}[t]
\centering
\includegraphics[width=\textwidth]{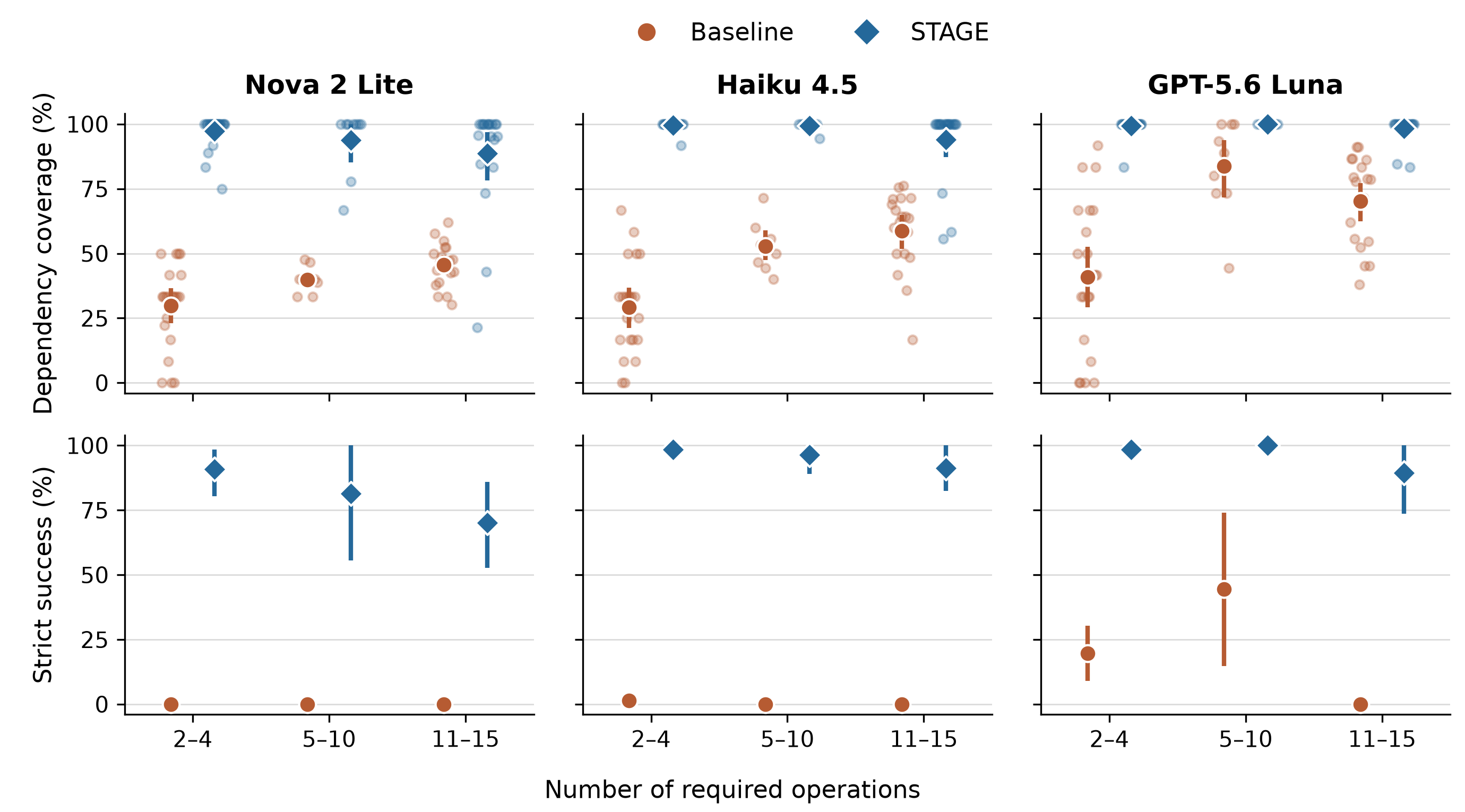}
\caption{Dependency coverage above and strict success below on Smart Dispute by required procedure length. Small points in the coverage panels show case means over evaluable runs. Large markers in both rows show equally weighted case averages with 95\% case-bootstrap intervals. Strict success is a per-run rate, distinct from $\mathrm{Pass}^{3}$.}
\label{fig:rq3-chain-length}
\end{figure*}

\subsection{RQ2: Fixed-Graph Component Ablations}
\label{sec:fixed-graph-ablation-results}

\begin{table}[t]
\centering
\small
\setlength{\tabcolsep}{2pt}
\renewcommand{\arraystretch}{1.08}
\begin{tabular*}{\columnwidth}{@{\extracolsep{\fill}}llccc@{}}
\toprule
\textbf{Benchmark}
& \textbf{Model}
& \shortstack{\textbf{Full-policy}\\\textbf{nodes}}
& \shortstack{\textbf{LLM}\\\textbf{routing}}
& \textbf{\textsc{Stage}} \\
\midrule
Telecom & GPT-5.6 Luna & 70.0 & 45.0 & \textbf{92.5} \\
 & & (-22.5) & (-47.5) & \\
 & Haiku 4.5 & 25.0 & 20.0 & \textbf{72.5} \\
 & & (-47.5) & (-52.5) & \\
\midrule
Smart Dispute & GPT-5.6 Luna & 64.0 & 84.0 & \textbf{96.0} \\
 & & (-32.0) & (-12.0) & \\
 & Haiku 4.5 & 16.0 & 34.0 & \textbf{90.0} \\
 & & (-74.0) & (-56.0) & \\
\bottomrule
\end{tabular*}
\caption{Fixed-graph ablations for \(\mathrm{Pass}^{3}\) (\%). Values below the scores show differences from \textsc{Stage} in percentage points.}
\label{tab:fixed_graph_ablation}
\end{table}

Table~\ref{tab:fixed_graph_ablation} examines the contributions of policy localization and routing ownership using fixed execution graphs. \textsc{Stage} achieves the highest \(\mathrm{Pass}^{3}\) in all four benchmark--model comparisons.

\paragraph{Policy localization.}
Replacing node-scoped context with the complete policy reduces \(\mathrm{Pass}^{3}\) by 22.5--74.0 percentage points across the four comparisons. These reductions occur with the execution graph and coordinator unchanged, demonstrating the contribution of policy localization to repeated-run success. One possible explanation is that the full policy imposes context demands beyond the effective processing capacity of the comparatively economical models evaluated here, making relevant conditions harder to identify and apply. The ablation does not isolate model capacity as the cause, but supports supplying each node with the policy context relevant to its assigned task.

\paragraph{Routing ownership.}
Allowing the model to select the next node from the full graph reduces \(\mathrm{Pass}^{3}\) by 12.0--56.0 percentage points across the four comparisons, while retaining node-scoped policy context. This introduces additional decision pressure, as the model must reason about workflow progress and select among graph-wide alternatives alongside completing its local task. The consistent reductions support separating local task reasoning from procedural control, allowing the model to focus on its assigned task while the coordinator determines the next permitted transition.

\subsection{RQ3: Procedural Compliance by Required Procedure Length}
\label{sec:rq3-complexity}

Figure~\ref{fig:rq3-chain-length} compares dependency coverage and strict success across required procedure lengths in Smart Dispute. \textsc{Stage} achieves higher dependency coverage in all nine model--length comparisons. On long procedures, coverage reaches 88.7--98.3\%, compared with 45.6--70.2\% for the baseline. However, baseline strict success is 0\% for all three models in this group, indicating that substantial partial coverage does not translate into successful execution of all procedural requirements. \textsc{Stage} achieves 70.2--91.2\% strict success on the same cases.

Across all three models, \textsc{Stage}'s strict success is lower on long procedures than on short procedures, although the extent of the decline varies. It nevertheless remains substantially higher than the baseline.

These findings are consistent with the intended division of responsibility in \textsc{Stage}. Localized policy context supports reasoning within each task, while shared state and coordinator-owned progression preserve procedural obligations across tasks. The fixed-graph ablations provide complementary evidence for these design choices. Appendix~\ref{sec:smart-dispute-analysis} provides detailed metric definitions, the complementary outcome-safety results, and a step-level analysis of required operations matched or missed across runs.

\section{Related Work}
\label{sec:related-work}

Policy-enforcement systems constrain agent behavior through checks around execution. ToolGuards compiles policy documents into guard code for tool calls \citep{zwerdling2025policyguards}, while AgentSpec expresses runtime rules through triggers, predicates, and enforcement actions \citep{wang2025agentspec}. PolicyGuide uses a compiled policy graph and a proactive verifier to track progress and provide step-specific remediation \citep{kang2026policyguide}. In \textsc{Stage}, the coordinator executes the policy graph by invoking bounded tasks, validating their results, and applying declared transitions.

Workflow compilation is the closest line of work. ARTIC makes data dependencies explicit through artifact reads and writes, constraint checks, and control transfers \citep{xu2026artic}. Compile, Then Page combines compiled SOP programs with active-frame paging \citep{yu2026compilepage}, and COVENANT interprets workflow control-flow graphs with checks before committing state or advancing \citep{wang2026covenant}. \textsc{Stage} builds on these ideas through an integrated runtime with typed task contracts, shared execution memory, controlled evidence access, and deterministic coordination. Our study examines policy localization and routing ownership on fixed graphs, evaluates repeated-run reliability across public and banking workflows, and analyzes procedural compliance across required procedure lengths in Smart Dispute.

\section{Conclusion}
\label{sec:conclusion}

We presented \textsc{Stage}, an integrated execution framework combining typed task nodes, shared execution memory, document tools, and deterministic orchestration for enterprise workflows. Across three public benchmarks and two proprietary banking workflows, \textsc{Stage} improves task success and repeated-run reliability over the baseline in all 18 benchmark--model comparisons. Fixed-graph ablations demonstrate the contributions of localized policy context and coordinator-owned routing. The Smart Dispute analysis further shows improved coverage of required operations across procedure lengths, including long workflows.

These findings show that execution design can improve reliable workflow completion while keeping the underlying model fixed. \textsc{Stage} bounds local reasoning, preserves information across tasks, and assigns procedural progression to the runtime, reducing the coordination responsibilities placed on the model. The results support an approach to enterprise agent design in which local judgments are connected through shared state and explicit execution rules.

\section*{Limitations}
\label{sec:limitations}

Our primary evaluation covers five workflows, with four models on the public benchmarks and three on the proprietary banking workflows. The results may not generalize to other policies, domains, models, or distribution shifts. Restricted access to the proprietary policies and case data used for Smart Dispute and Paystub Fraud Detection limits independent reproduction of these evaluations. The three public benchmarks provide externally accessible evaluation settings.

\textsc{Stage} assumes that each frozen graph faithfully represents the governing policy. Construction or validation errors can systematically enforce an incorrect procedure. Procedural conformance also does not guarantee correct semantic judgment, as an execution can follow the permitted transitions while making an incorrect local interpretation.

We do not yet quantify graph-authoring effort, validation cost, or maintenance under policy changes. Cost estimates depend on provider pricing and cache reuse. We do not provide a controlled latency comparison.

\section*{Ethical Considerations}

The proprietary banking cases used in this study are synthetic, de-identified, and fully masked. No confidential or private information is available to researchers or will be disclosed. The datasets remain subject to intellectual-property and privacy protections.

\bibliography{custom}
\clearpage

\appendix

\section{Policy-to-graph Construction}
\label{sec:policy2graph-construction}

\begin{figure}[t]
\centering
\includegraphics[width=\columnwidth]{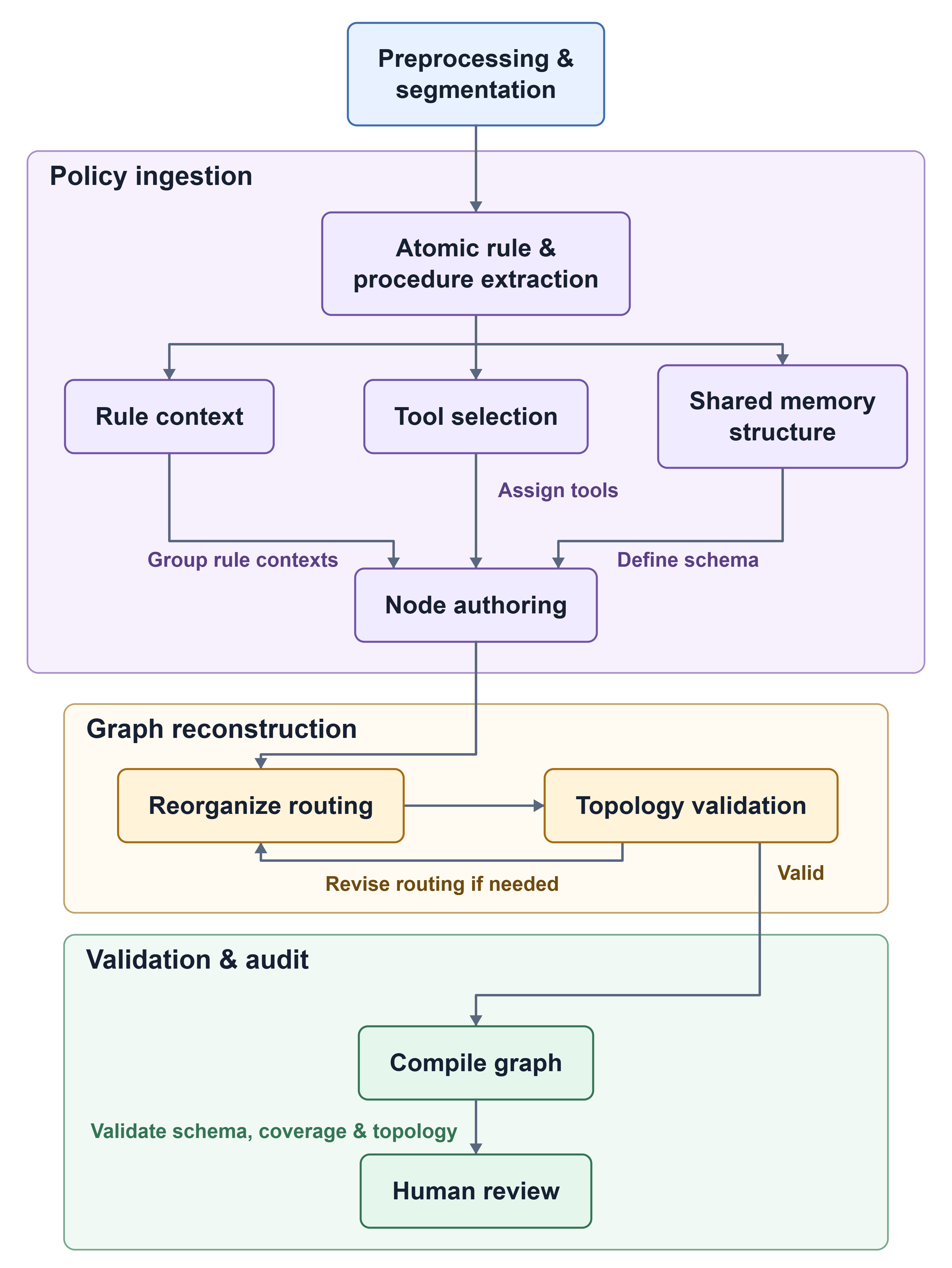}
\caption{Offline policy-to-graph construction. Extracted rules and procedures inform rule contexts, tool selection, and shared memory. These inputs support node authoring, followed by graph reconstruction, compilation and validation, and human review.}
\label{fig:policy2graph-construction}
\end{figure}

Figure~\ref{fig:policy2graph-construction} summarizes the offline construction workflow. After segmentation, rule and procedure extraction identifies atomic rules, atomic actions, and routing, and establishes links among rules, actions, and terminology. These outputs inform rule contexts, tool requirements matched against the available registry, and a shared memory structure. Node authoring groups rule contexts according to model capability, assigns registered tools, and specifies structured output schemas. Graph reconstruction reorganizes routing and iteratively validates topology. The resulting graph is compiled and validated for schema, policy coverage, and topology before human review, approval, and freezing.

The evaluated graphs were generated with Claude Opus 4.8. The construction prompts, implementation code, and internal state schemas are proprietary artifacts used in the internal policy-to-graph pipeline and cannot be released because of intellectual-property restrictions.

\section{Agent Inference Cost}
\label{sec:inference-cost}

Table~\ref{tab:inference-cost} reports mean agent inference cost per case in USD, averaged over three runs, alongside the corresponding gain in \(\mathrm{Pass}^{3}\). Costs cover available agent usage records and exclude simulator and evaluator charges.

Two sources of overhead accompany \textsc{Stage}'s execution structure. First, each node produces structured findings, evidence references, and state updates for compliance checks and review, adding output requirements beyond those of the baseline. Second, coordinator-triggered retries after invalid results repeat prompt processing and generation. Additional turns for tool use or evidence retrieval can further increase cost. Inference cost therefore depends on how readily a model satisfies the task and result contract, as well as the number of executed nodes.

The observed costs vary substantially by model and workflow. Claude Haiku 4.5 has the highest \textsc{Stage} cost in all five evaluated workflows, at 5.3--14.0 times its baseline cost. Gemma 4 26B A4B remains closer to its baseline across the three public workflows, with cost ratios of 0.77--1.34. These patterns are consistent with differences in model capability and suitability for the task, but aggregate costs do not isolate the effects of output requirements, retries, tool use, and pricing. Within the fixed-graph ablations, localized policy context reduces cost by 25.5--82.7\% relative to full-policy nodes, showing how policy scope can moderate the overhead of structured execution.

\begin{table*}[t]
\centering
\small
\setlength{\tabcolsep}{4pt}
\renewcommand{\arraystretch}{1.08}
\begin{tabular*}{\textwidth}{@{\extracolsep{\fill}}llccccc@{}}
\toprule
\textbf{Benchmark} & \textbf{Model}
& \textbf{Baseline} & \textbf{\textsc{Stage}} & \shortstack{\(\Delta\mathrm{Pass}^{3}\)\\(pp)} & \textbf{Full-policy} & \textbf{LLM-routing} \\
\midrule
Referral Abuse & GPT-5.6 Luna & 0.0017 & 0.0201 & \(+7.5\) & --- & --- \\
& Claude Haiku 4.5 & 0.0092 & 0.1287 & \(+9.5\) & --- & --- \\
& Amazon Nova 2 Lite & 0.0190 & 0.0531 & \(+16.0\) & --- & --- \\
& Gemma 4 26B A4B & 0.0074 & 0.0099 & \(+10.0\) & --- & --- \\
\midrule
Retail & GPT-5.6 Luna & 0.0046 & 0.0202 & \(+27.5\) & --- & --- \\
& Claude Haiku 4.5 & 0.0251 & 0.1320 & \(+45.0\) & --- & --- \\
& Amazon Nova 2 Lite & 0.0232 & 0.0238 & \(+32.5\) & --- & --- \\
& Gemma 4 26B A4B & 0.0173 & 0.0134 & \(+10.0\) & --- & --- \\
\midrule
Telecom & GPT-5.6 Luna & 0.0089 & 0.0213 & \(+60.0\) & 0.1230 & 0.0528 \\
& Claude Haiku 4.5 & 0.0449 & 0.3185 & \(+47.5\) & 0.4277 & 0.1930 \\
& Amazon Nova 2 Lite & 0.0327 & 0.0660 & \(+30.0\) & --- & --- \\
& Gemma 4 26B A4B & 0.0228 & 0.0229 & \(+52.5\) & --- & --- \\
\midrule
Smart Dispute & GPT-5.6 Luna & 0.0067 & 0.0272 & \(+42.0\) & 0.1208 & 0.0833 \\
& Claude Haiku 4.5 & 0.0362 & 0.2090 & \(+66.0\) & 0.6329 & 0.3984 \\
& Amazon Nova 2 Lite & 0.0703 & 0.0564 & \(+46.0\) & --- & --- \\
\midrule
Paystub Fraud Detection & GPT-5.6 Luna & 0.0094 & 0.0415 & \(+47.1\) & --- & --- \\
& Claude Haiku 4.5 & 0.0493 & 0.3781 & \(+64.7\) & --- & --- \\
& Amazon Nova 2 Lite & 0.0172 & 0.1251 & \(+35.3\) & --- & --- \\
\bottomrule
\end{tabular*}
\caption{Mean agent inference cost per case across three runs. Cost columns are in USD. \(\Delta\mathrm{Pass}^{3}\) is the percentage-point gain of \textsc{Stage} over the baseline from Table~\ref{tab:success-reliability}. Full-policy and LLM-routing are the policy-localization and routing-ownership ablations. Dashes indicate settings not evaluated.}
\label{tab:inference-cost}
\end{table*}

\section{Reconstructed Benchmark Graphs}
\label{sec:benchmark-graphs}

\begin{figure*}[t]
\centering
\begin{minipage}[t]{0.17\textwidth}
\centering
\includegraphics[height=0.11\textheight,keepaspectratio]{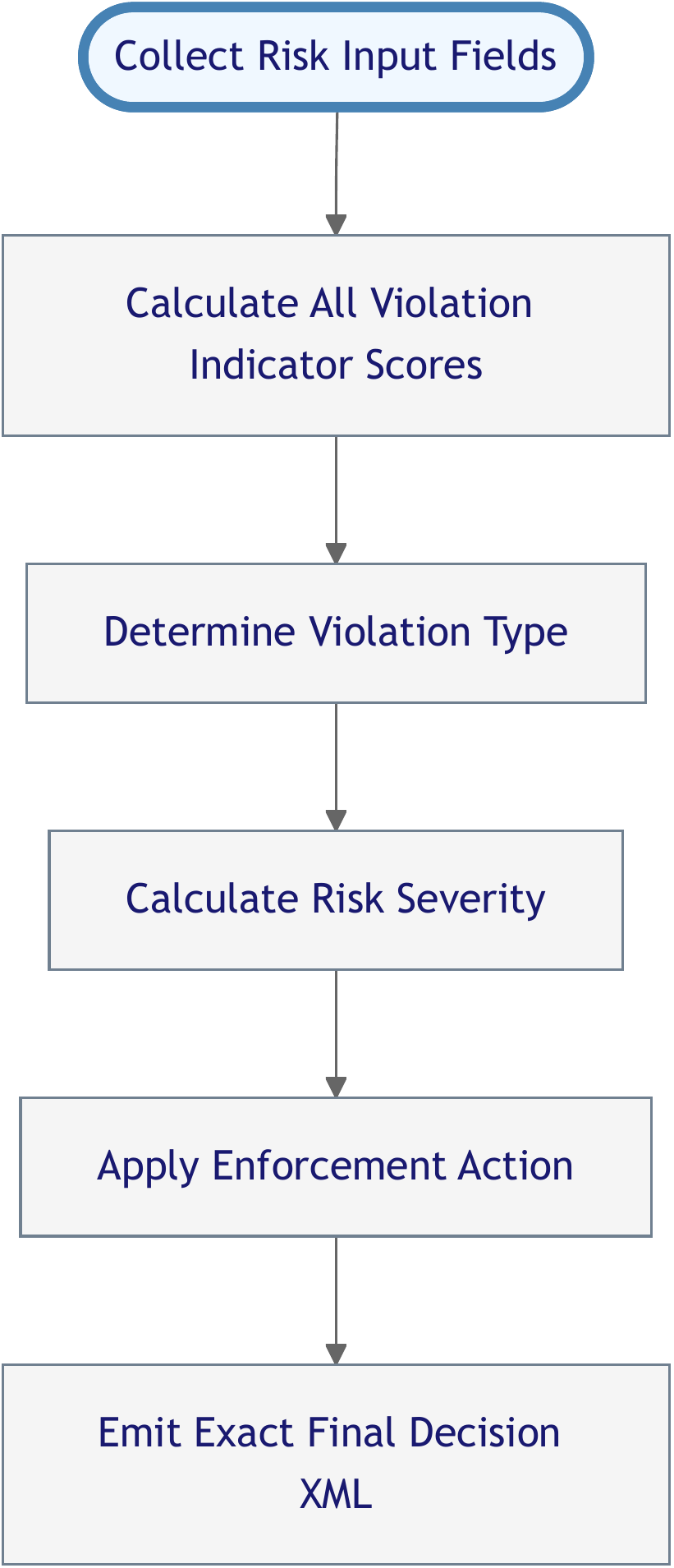}
\par\smallskip
{\fontsize{8.5pt}{10pt}\selectfont\textbf{(a) Referral Abuse}\par}
\end{minipage}
\hfill
\begin{minipage}[t]{0.30\textwidth}
\centering
\includegraphics[height=0.11\textheight,keepaspectratio]{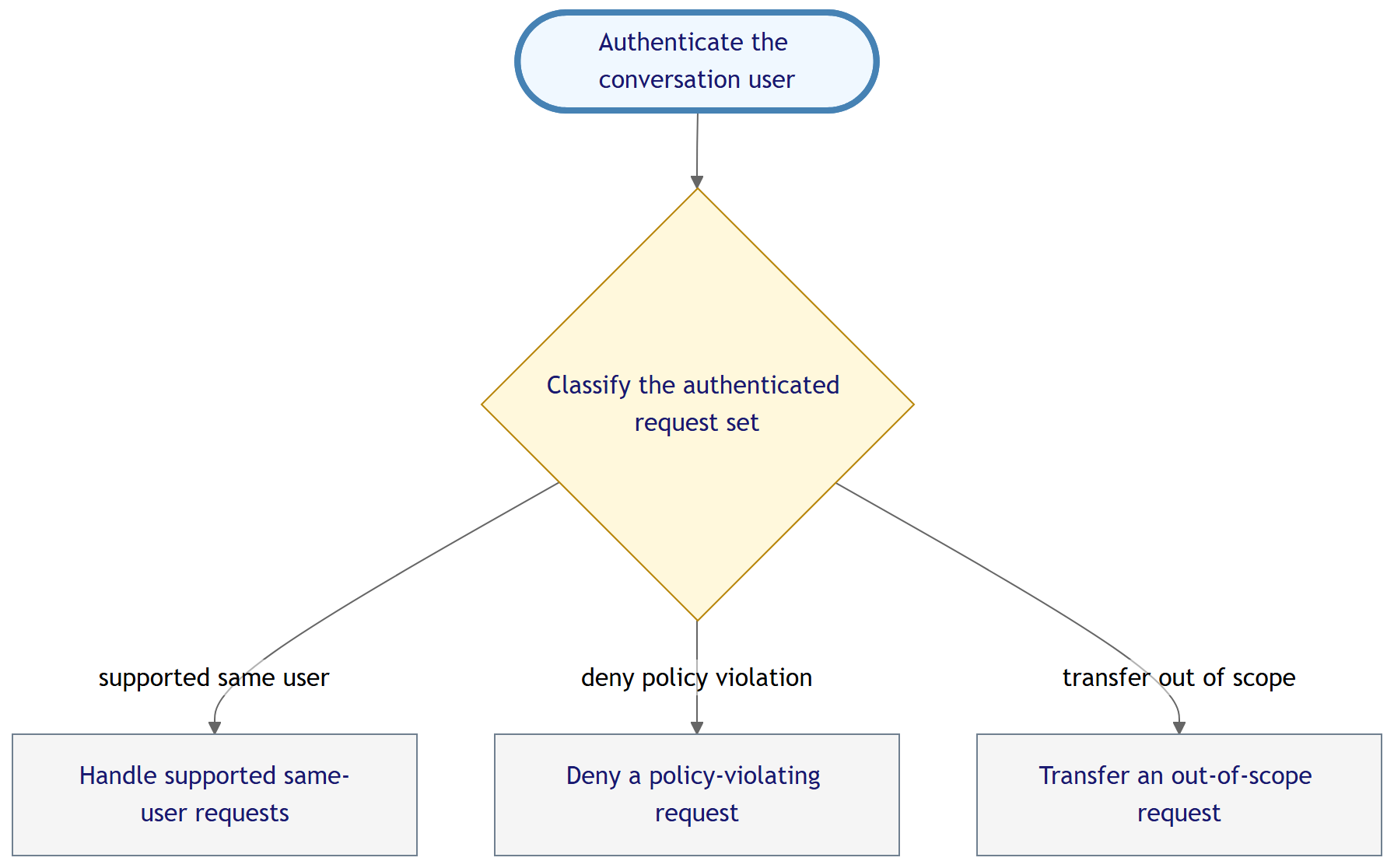}
\par\smallskip
{\fontsize{8.5pt}{10pt}\selectfont\textbf{(b) Retail}\par}
\end{minipage}
\hfill
\begin{minipage}[t]{0.51\textwidth}
\centering
\includegraphics[height=0.11\textheight,keepaspectratio]{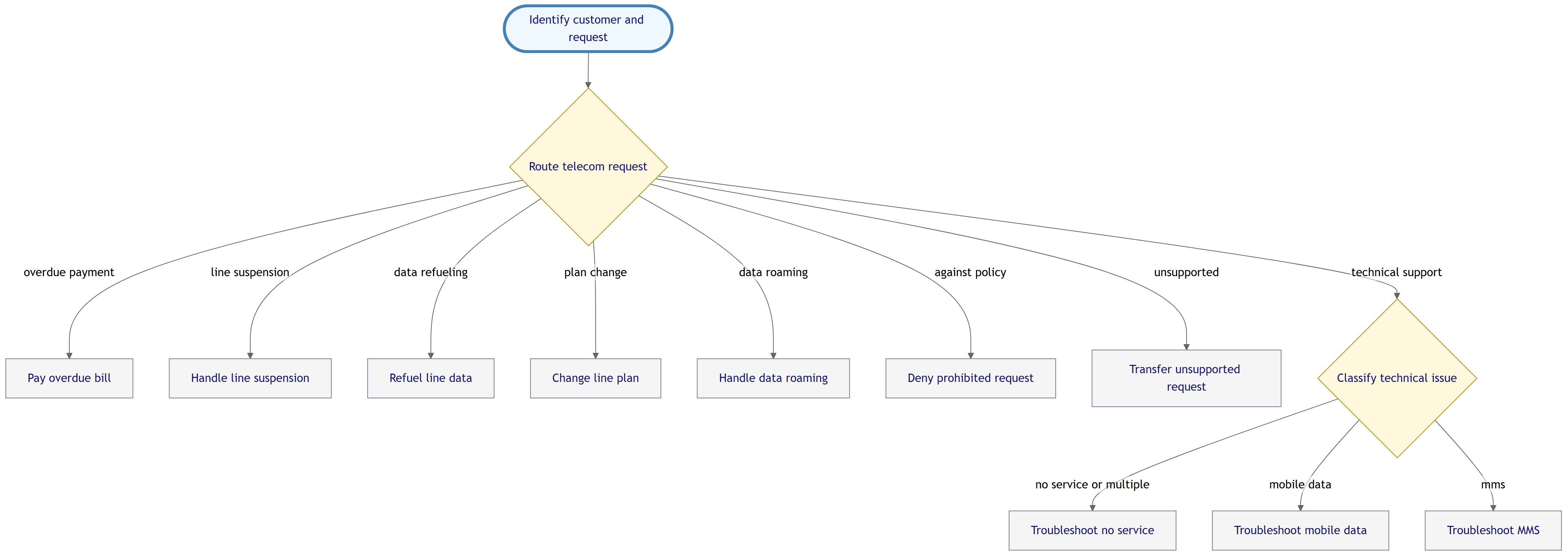}
\par\smallskip
{\fontsize{8.5pt}{10pt}\selectfont\textbf{(c) Telecom}\par}
\end{minipage}
\par\vspace{1.2em}
\begin{minipage}[t]{0.44\textwidth}
\centering
\includegraphics[width=\linewidth,height=0.66\textheight,keepaspectratio]{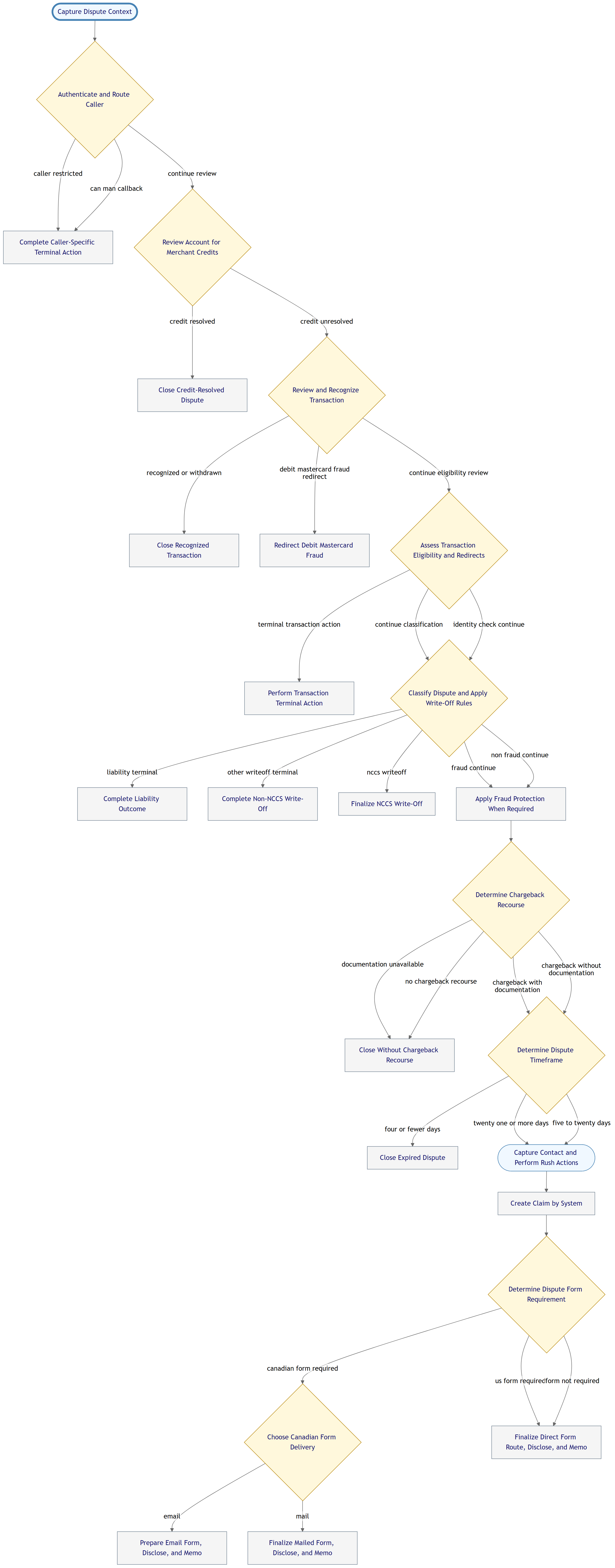}
\par\smallskip
{\fontsize{8.5pt}{10pt}\selectfont\textbf{(d) Smart Dispute}\par}
\end{minipage}
\hfill
\begin{minipage}[t]{0.54\textwidth}
\centering
\includegraphics[width=\linewidth,height=0.66\textheight,keepaspectratio]{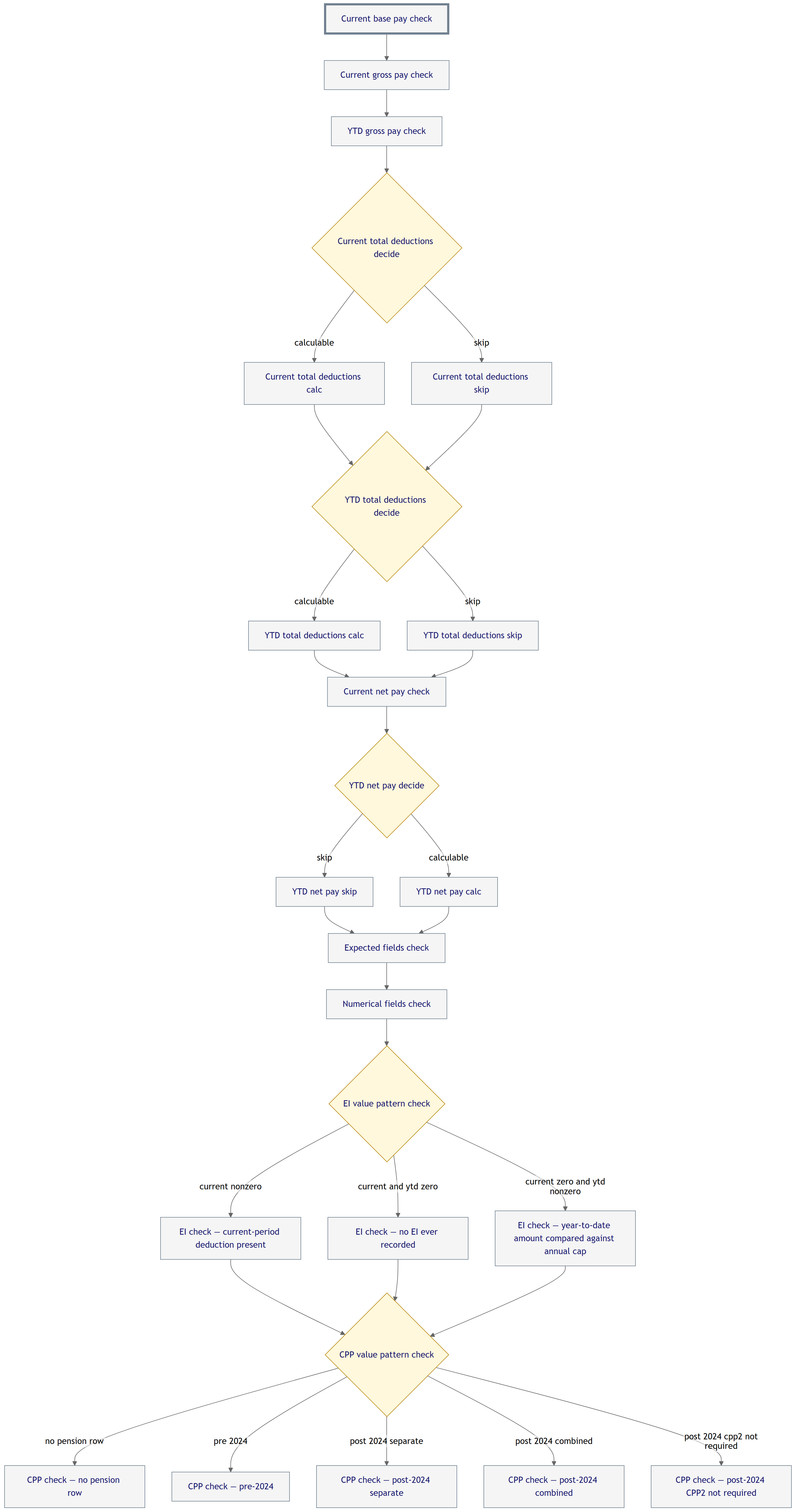}
\par\smallskip
{\fontsize{8.5pt}{10pt}\selectfont\textbf{(e) Paystub Fraud Detection}\par}
\end{minipage}
\caption{Reconstructed benchmark execution graphs. Referral Abuse is sequential, Retail provides the shallowest routing structure, Telecom adds broader request routing and a nested technical-support decision, and Smart Dispute contains the deepest sequence and most conditional structure. Paystub Fraud Detection combines pay consistency checks with conditional deduction checks. The panels include only non-sensitive node titles and outcome labels.}
\label{fig:benchmark-graphs-combined}
\end{figure*}

Figure~\ref{fig:benchmark-graphs-combined} shows the reviewed execution graphs used in the evaluation. SOP-Bench Referral Abuse contains 6 sequential stages for risk calculation, violation classification, severity assessment, and enforcement. Retail contains 5 nodes and 1 decision, providing a compact routing structure. Telecom contains 13 nodes and 2 decisions, adding wider request routing and a deeper technical-support branch. Smart Dispute contains 26 nodes and 9 decisions, making it the deepest and most highly branched graph. Paystub Fraud Detection contains 25 nodes and 5 decisions, combining pay consistency checks with conditional deduction checks.

\section{Smart Dispute Metrics and Procedural Analysis}
\label{sec:smart-dispute-analysis}

Smart Dispute is a proprietary banking benchmark for handling customer disputes over card transactions according to an internal banking policy. The agent interacts with a simulated customer and uses business tools to complete the checks and actions required for each case.

\textbf{Outcome-safety success} requires completed execution, the correct terminal outcome, and no forbidden action attempts. It defines the case-level passes used for Smart Dispute's \(\mathrm{Pass}^{1}\) and \(\mathrm{Pass}^{3}\). \textbf{Strict success} additionally requires every required operation to match under the evaluator's dependency checks. It measures procedural completeness and outcome safety within one run.

\textbf{Dependency coverage} measures the fraction of required operations matched in a run. For case $i$ and run $r$, let $L_i$ be the number of required operations and $M_{i,r}$ the number matched. Then
\begin{equation}
C_{i,r}=\frac{M_{i,r}}{L_i}.
\label{eq:dispute-dependency-coverage}
\end{equation}

An operation matches when a successful event for it follows a successful event for each direct prerequisite. \textbf{Step coverage} is the fraction of evaluable runs in which an individual required operation matches. Both coverage measures describe final matching.

\begin{figure*}[t]
\centering
\includegraphics[width=\textwidth]{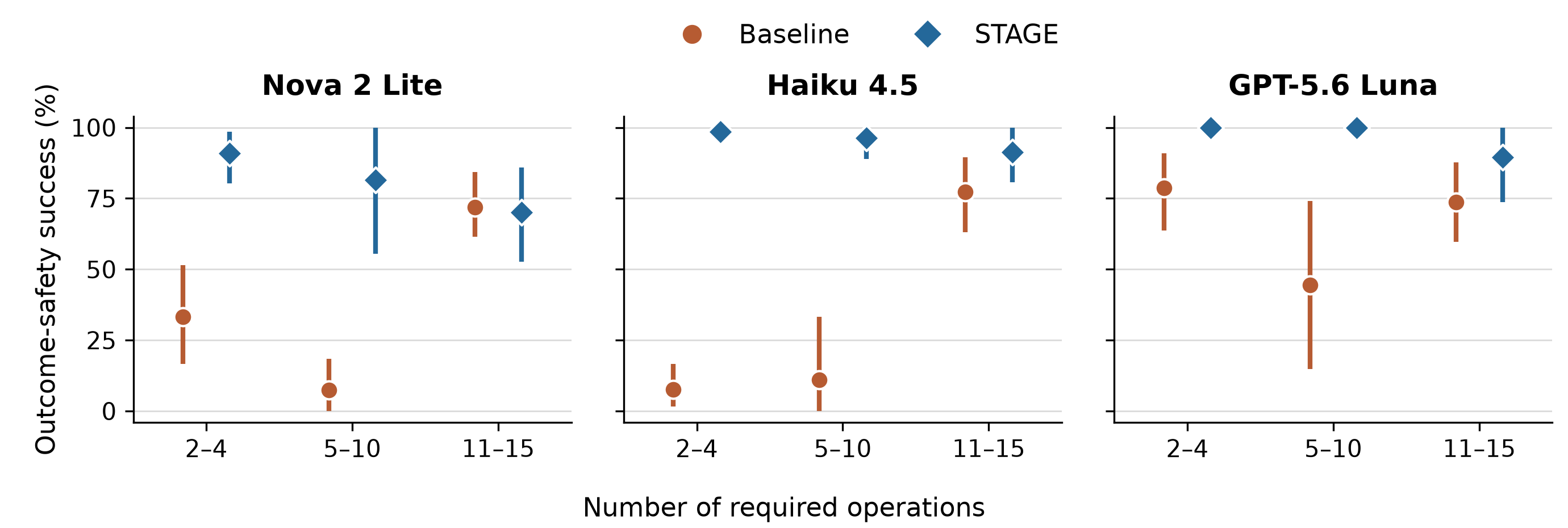}
\caption{Outcome-safety success on Smart Dispute by required procedure length. Points show equally weighted case averages over three runs, with 95\% case-bootstrap intervals. Outcome-safety success requires completed execution, the correct terminal outcome, and no forbidden action attempts. These are per-run success rates, distinct from $\mathrm{Pass}^{3}$.}
\label{fig:dispute-outcome-strict}
\end{figure*}

\begin{figure*}[t]
\centering
\includegraphics[width=\textwidth]{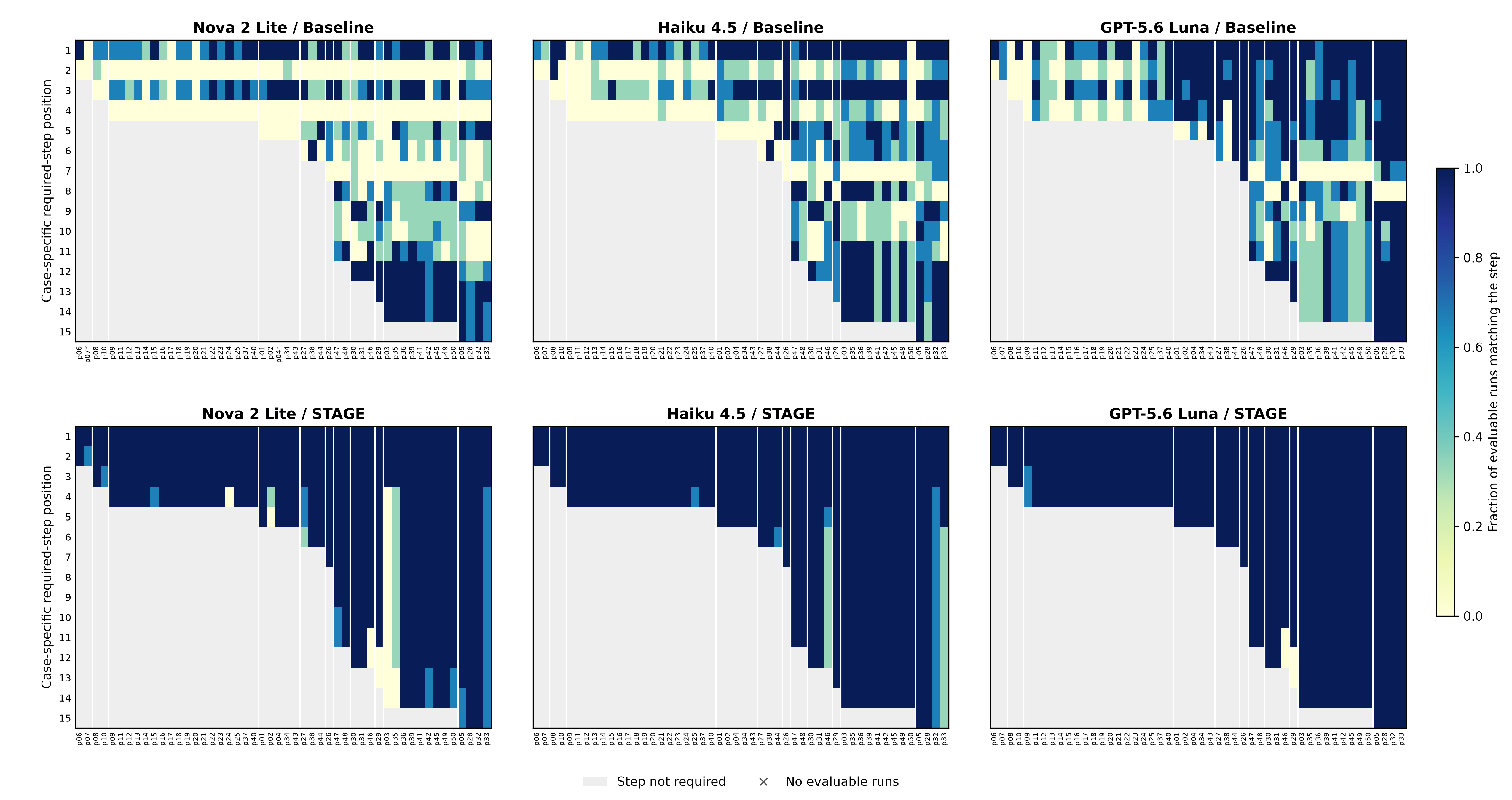}
\caption{Step coverage across 50 Smart Dispute cases, ordered by required procedure length. Rows index case-specific required operations. Color shows the fraction of evaluable runs matching each operation. Gray denotes unrequired positions, an asterisk marks fewer than three evaluable runs, and $\times$ denotes no evaluable runs. The heatmap shows final matching, not first-error timing.}
\label{fig:dispute-step-coverage}
\end{figure*}

Figure~\ref{fig:dispute-outcome-strict} reports outcome-safety success across required procedure lengths, complementing the dependency coverage and strict success in Figure~\ref{fig:rq3-chain-length}. On long procedures, baseline outcome-safety success reaches 71.9--77.2\% across models, while its strict success in Figure~\ref{fig:rq3-chain-length} is 0\% for every model. \textsc{Stage} achieves 70.2--91.2\% strict success, showing that correct outcomes can be accompanied by complete execution of the required procedure under the evaluator's dependency checks. Figure~\ref{fig:dispute-step-coverage} provides the corresponding step-level view. The baseline repeatedly misses required operations, while \textsc{Stage} matches most requirements consistently.

\end{document}